\documentclass{article}

\usepackage{times}
\usepackage{graphicx} 
\usepackage{subfigure} 

\usepackage{natbib}

\usepackage{algorithm}
\usepackage{algorithmic}

\usepackage{hyperref}

\usepackage[accepted]{implicitworkshop2017}

\icmltitlerunning{Can GAN Learn Topological Features of a Graph?}

\begin{document} 

\twocolumn[
\icmltitle{Can GAN Learn Topological Features of a Graph?}



\begin{icmlauthorlist}
\icmlauthor{Weiyi Liu}{to,goo}
\icmlauthor{Pin-Yu Chen}{goo}
\icmlauthor{Hal Cooper}{ed}
\icmlauthor{Min Hwan Oh}{ed}
\icmlauthor{Sailung Yeung}{bos}
\icmlauthor{Toyotaro Suzumura}{goo}
\end{icmlauthorlist}

\icmlaffiliation{to}{University of Electronic Science and Technology of China, Chengdu, China}
\icmlaffiliation{goo}{IBM Watson Research Center, Yorktown Heights, New York, USA}
\icmlaffiliation{ed}{Columbia University, 116th St Broadway, New York, USA}
\icmlaffiliation{bos}{Boston University, Boston, USA}

\icmlcorrespondingauthor{Weiyi Liu}{weiyiliu@us.ibm.com}

\icmlkeywords{Generative Adversarial Network, Community Detection, Topology Feature Extraction}

\vskip 0.3in
]



\printAffiliationsAndNotice{}

\begin{abstract}
This paper is first-line research expanding GANs into graph topology analysis. By leveraging the hierarchical connectivity structure of a graph, we have demonstrated that generative adversarial networks (GANs) can successfully capture topological features of any arbitrary graph, and rank edge sets by different stages according to their contribution to topology reconstruction. Moreover, in addition to acting as an indicator of graph reconstruction, we find that these stages can also preserve important topological features in a graph.
\end{abstract} 

\section{Introduction}
\label{Intro}
With the rise of social networking and the increase of data volume, graph topology analysis has become an active research topic in analyzing structured data. Many graph topology analysis tools have been proposed to tackle particular kinds of topology discovered in real life. 

One common problem of existing graph analysis tools is that these methods are highly sensitive to a presumed topology of a graph and hence suffer from model mismatch. For example, BA\cite{barabasi1999emergence} model are capable of capturing scale-free features of a graph, WS model\cite{watts1998collective} is suitable for depicting small-world feature of a graph. Modularity based community detection methods\cite{xiang2016local} are suitable for a graph consisting of non-overlapping communities, while link based community detection methods\cite{delis2016scalable} perform well on a graph with highly overlapping communities. Another problem that arises from real-world observations is that a graph of interest is often a mixture of different types of topological models, and no topological model so far can fit well to all kinds of real-life graphs. For example, an online social network always have both scale-free and small-world features. But typical BA or WS graph model fails to capture these two features at the same time. What's more, a synthetic graph from WS network cannot have the scale-free features, and vice versa. In addition, for community detection methods, as one cannot decide the topological features of a graph in the first place (e.g., overlapping communities or not, balanced community or not, deep community or not\cite{chen2015deep}), it is hard to tell which community detection methods should we use to uncover rightful communities.

In general, the reason why these problems happen is that in graph analysis area, we lack a general model-free tool which can automatically capture important topological features of any arbitrary graph. But fortunately, in image processing area, generative adversarial networks (GANs) have been widely used to capture features of an image\cite{goodfellow2014generative,goodfellow2016nips,denton2015deep,chen2016infogan,zhao2016energy,nowozin2016f}. In this position paper, we expand the use of GANs into graph topology analysis area, and propose a Graph Topology Interpolator (GTI) method to automatically capture topological features of any kinds of a graph. To the best of our knowledge, this paper is the first paper to introduce GAN into graph topology analysis area. 

With the help of GANs, GTI can automatically capture important topological features of any kinds of a graph, and thereby overcoming the ``one model cannot fit all'' issue. What's more, unlike any convolutional neural network (CNN) related graph analysis tools\cite{bruna2014spectral,henaff2015deep,duvenaud2015convolutional,radford2015unsupervised,defferrard2016convolutional,kipf2016semi} focusing mainly on feature extraction, GTI also has the ability to reconstruct a weighted adjacency matrix of the graph, where different weights in the matrix indicates the level of contribution of edges to the entire topology of the original graph. By ranking edges with different weights into an ordered stages, these stages not only reveal the reconstruction process of a graph, but also can be used as an indicator of the importance of  topological features in a reconstruction process. In summary, by analyzing these stages, GTI provides a way to accurately capture important topological features of a single graph of arbitrary structure.

\section{Graph Topology Interpolator (GTI)}
\begin{figure}[ht]
\vskip 0.2in
\begin{center}
\centerline{\includegraphics[width=\columnwidth]{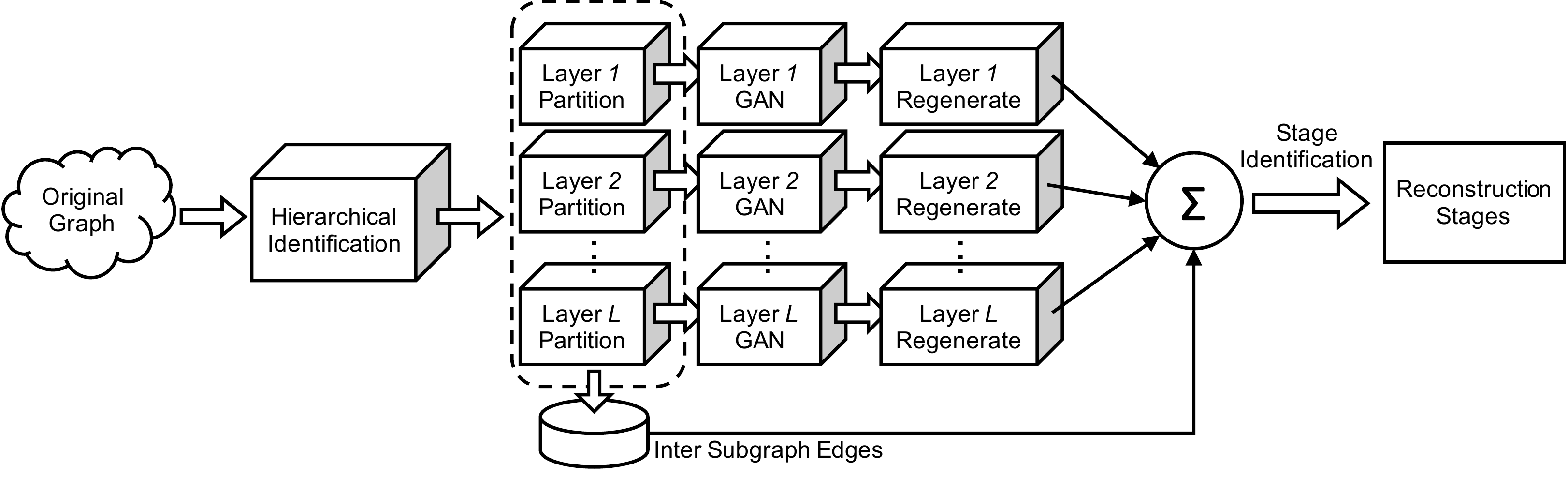}}
\caption{Workflow for graph topology interpolater (GTI).}
\label{fig-HGG-flow}
\end{center}
\vskip -0.2in
\end{figure}

In this section, we demonstrate the workflow of the Graph Topology Interpolator (GTI) (Figure \ref{fig-HGG-flow}). Overall, GTI takes a graph as an input, constructs hierarchical layers, trains a GAN for each layer, combines outputs from all layers to identify reconstruction stages of the original graph automatically. Specifically, GTI produces stages (a set of edges) of the original graph, where these stages not only have the ability to capture the important topological features of the original graph but also can be interpreted as steps for graph reconstruction process.
In the rest of this section, we give a brief introduction for each module.

\textbf{Hierarchical Identification Module}: 
By leveraging Louvain hierarchical community detection method\cite{blondel2008fast}, this module identifies hierarchical layers of the original graph. For each layer, the number of communities in the layer works as a criterion for how many subgraphs a layer should pass to the next module.

\textbf{Layer Partition Module}: Although Louvain has the ability to identify communities for a layer, we cannot constrain the size of any community, which is hard for a convolutional neural network with fully connected layers to capture features.
Instead, we introduce METIS graph partition tool \cite{karypis1995metis} to identify non-overlapping subgraphs within this layer, while the number of subgraphs equals to the number of communities in the layer.

\textbf{Layer GAN Module}: As different layers present different topological features of the original graph in each hierarchy, rather than directly using one GAN to learn the whole graph, we use different GANs to learn features for each layer separately. 
For each GAN upon each layer, the generator is a deconvolutional neural networks with two fully connected layers and two deconvolutional layers, and the discriminator is a convolutional neural networks with two convolutional layers and two fully connected layers. 
For activation function, we use ``Leaky ReLu ($LR=max(x, 0.2\times x)$)'' instead of ``ReLu,'' as value $0$ has a specific meaning in adjacency matrix (i.e., absence of edges). 
What's more, we replace ``Max Pooling'' layer with ``Batch normalization'' layer, as the former only selects the maximum value in the feature map and ignores other values, but the latter will synthesize all available information. 

By feeding adjacency matrices of all subgraphs in the layer into a GAN, and adopting the same loss function and optimization strategy (1000 iterations of ADAM\cite{kingma2014adam} with a learning rate of 0.0002) used in DCGAN \cite{radford2015unsupervised}, we find that the generator will eventually capture the important topological features of subgraphs in the corresponding layer, and is able to reproduce the weighted adjacency matrix of a subgraph in that layer.

\textbf{Layer Regenerate Module}: For a given layer with $M$ subgraphs of $k$ nodes, the corresponding generator trained by all subgraphs in the layer can regenerate the weighted adjacency matrix of a subgraph with $k$ nodes. Accordingly, by regenerating $M$ subgraphs, this module can reconstruct the weighted adjacency matrix of the layer. Please note that this reconstruction only restores edges within each non-overlapping subgraph, and does not include edges between subgraphs.

\textbf{All Layer Sum-up Module}: In this module, we use a linear function (see Equation \ref{eq-sum}) to aggregate weighted adjacency matrices of all layers together, along with the adjacency matrix of edges between subgraphs which we ignored in previous modules. The notation $re_G$ stands for the reconstructed weighted adjacency matrix for the original graph,
$G'_i, i \in L$ represents the reconstructed adjacency matrix for each layer, 
$E$ represents the adjacency matrix of inter subgraph edges, and $b$ represents a bias. Note that while each layer of the reconstruction may lose certain edge information, summing up the hierarchical layers along with $E$ will have the ability to reconstruct the entire graph.

\begin{equation}
re_G = \sum_{i=1}^L w_i G'_i + wE + b \label{eq-sum}
\end{equation}

To obtain the weight $w, w_i$ for each layer and the bias $b$, we introduce Equation \ref{eq-sum-loss} as the loss function, which is analogue to $KL$ divergence of two distributions (though of course $re_G$ and $G$ are not probability distributions). Here, we add $\epsilon=10^{-6}$ to avoid taking $log(0)$ or division by 0, and $vec(G)$ stands for vectorizing the weighted adjacency matrix $G$ with $N$ nodes.
By using 500 iterations of stochastic gradient descent (SGD) with learning rate 0.1 to minimize the loss function, this module outputs the optimized weighted adjacency matrix of the reconstructed graph. We then use these weights to identify the reconstruction stages for the original graph in the next module.

\begin{equation}
\textrm{ Loss }(re_{ G },G)=\sum _{ i\in \{1\cdots N^{ 2 }\} }^{  }{ vec(G+\epsilon )_i\cdot log\frac { vec(G+\epsilon )_i }{ vec(re_{ G }+\epsilon )_i }  }  
\label{eq-sum-loss}
\end{equation}

\textbf{Stage Identification Module}: 
Clearly, different edge weights in the obtained weighted adjacency matrix of the reconstructed graph $re_G$ from previous modules represent different degrees of contribution to the topology. Hence, we define an ordering of stages by decreasing weight, giving insight on how to reconstruct the original graph in terms of edge importance.
According to these weights, we can divide the network into several stages, with each stage representing a collection of edges greater than a certain weight. Here, we introduce the concept of a ``cut-value'' to turn $re_G$ into a binary adjacency matrix. As shown in Equation \ref{eq:stage_def}, We denote the $i$th largest unique weight-value as $CV_i$ (for ``cut value''), where $I[w\geq CV_i]$ is an indicator function for each weight being equal or larger than the $CV_i$.
\begin{equation}
re_G^i = re_G I[w\geq CV_i]
\label{eq:stage_def}
\end{equation}

\section{Evaluation}
To show the stages GTI identifies have the ability to capture topological features of a original graph, we use four synthetic and two real datasets to show that each stage preserves identifiable global (section \ref{global}) and local (section \ref{local}) topological features of the original graph during the graph reconstruction process. What's more, as each stage contains a subset of the original graphs edges, we can interpret each stage as a sub-sampling of the original graph. This allows us to compare with prominent graph sampling methodologies to emphasize our ability to retain important topological features (section \ref{graph-sampling}).

Table \ref{tab:Table-datasets} shows the detailed information for these datasets. 
All real datasets comes from Stanford Network Analysis Project (SNAP)\cite{SNAP}. 
All experiments in this paper were conducted locally on CPU using a Mac Book Pro with an Intel Core i7 2.5GHz processor and 16GB of 1600MHz RAM.

\begin{table}[t]
  \caption{Basic Graph Topology Information for Six Datasets}
  \label{tab:Table-datasets}
  \vskip 0.15in
  \begin{small}
  \centering
  \begin{tabular}{ccc|ccc}
  \hline
  \abovespace\belowspace
    Graphs 	& 	Nodes 	&	Edges & Graphs 	& 	Nodes 	&	Edges\\
    \hline
    \abovespace
    ER		&	500		&	25103 & Kron	&	2178	&	25103  \\
    BA		&	500		&	996	  & Facebook&	4039	&	88234\\
    \belowspace
    WS		&	500		&	500	  & Wiki-Vote&	7115	&	103689\\
    \hline
  \end{tabular}
 \end{small}
\vskip -0.1in
\end{table}

\subsection{Global Topological Features} \label{global}
\begin{figure}[ht]
  \subfigure[Degree distributions for ER and BA synthetic datasets]
  {
  	\label{syn1}
  \includegraphics[width=0.48\linewidth]{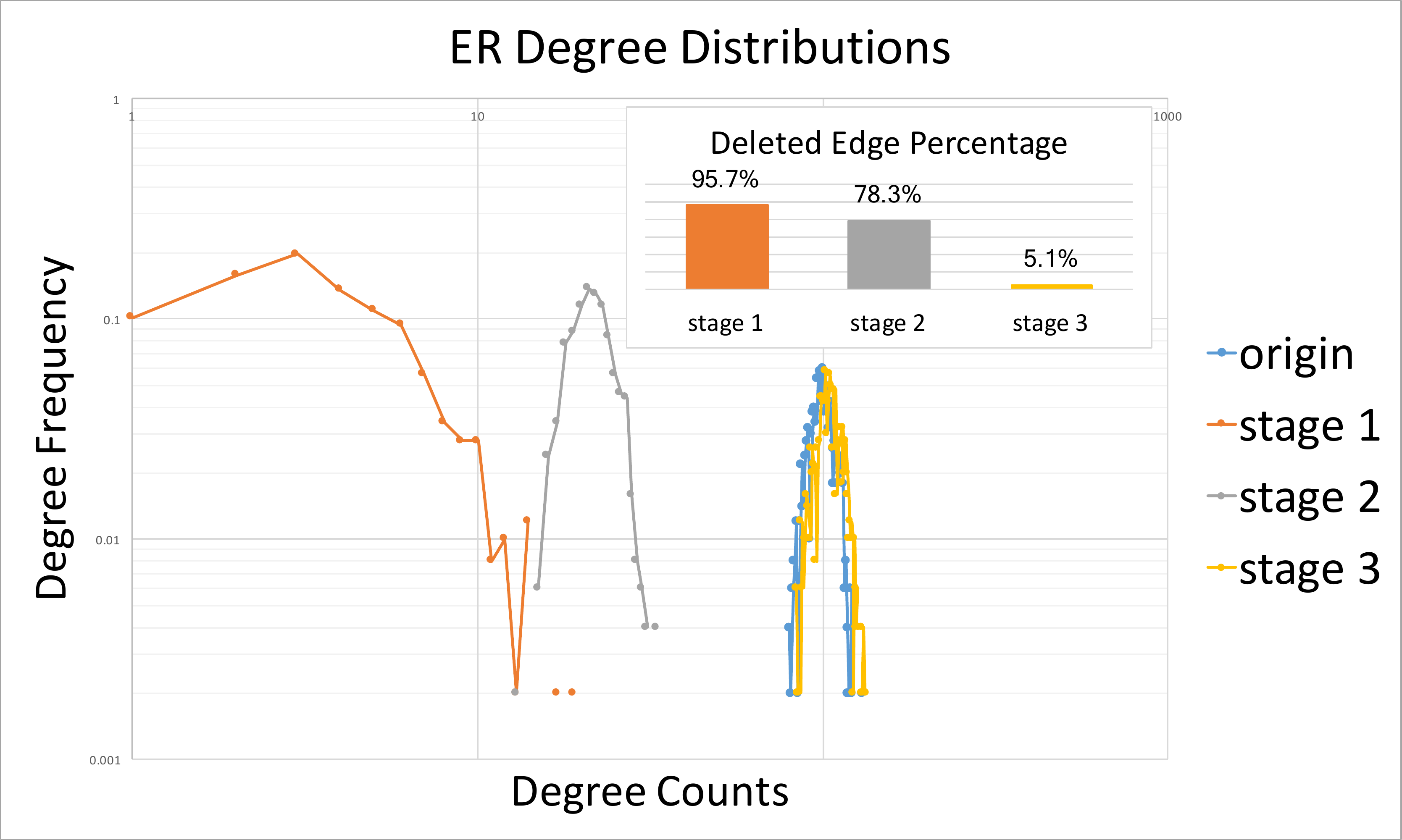}
  \includegraphics[width=0.48\linewidth]{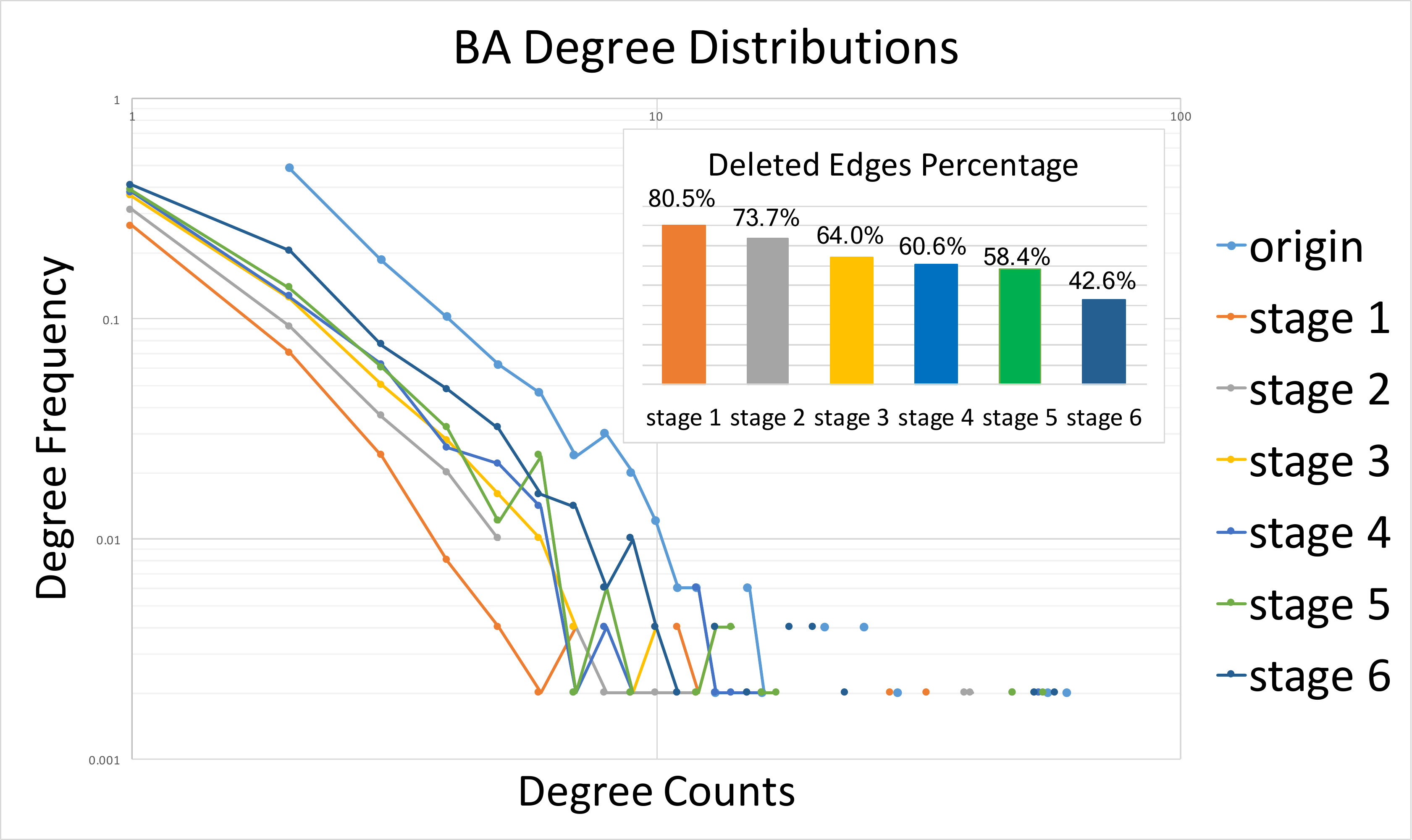}
  }
    \subfigure[Degree distributions for WS and Kron synthetic datasets]
  {
  	\label{syn2}
  \includegraphics[width=0.48\linewidth]{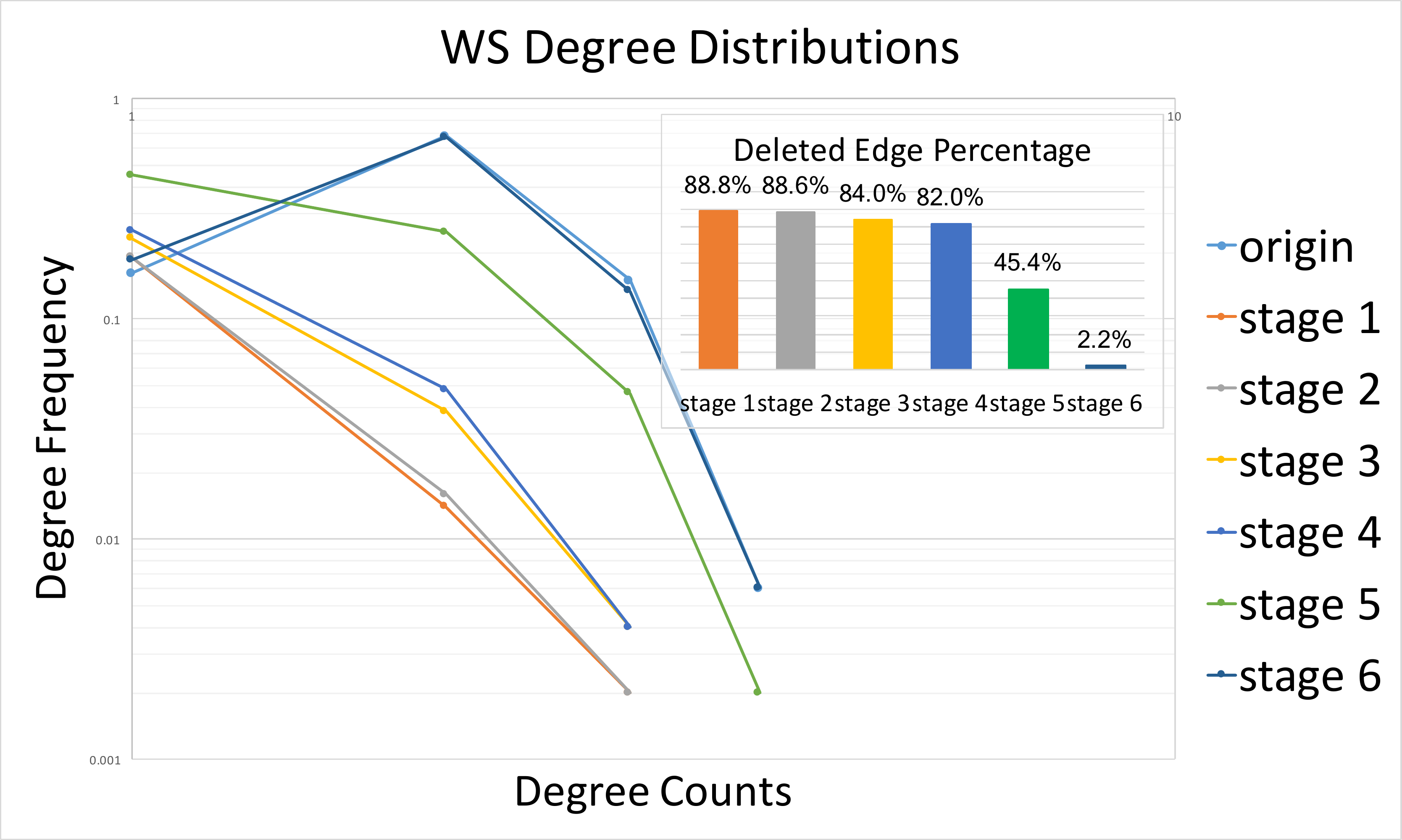}
  \includegraphics[width=0.48\linewidth]{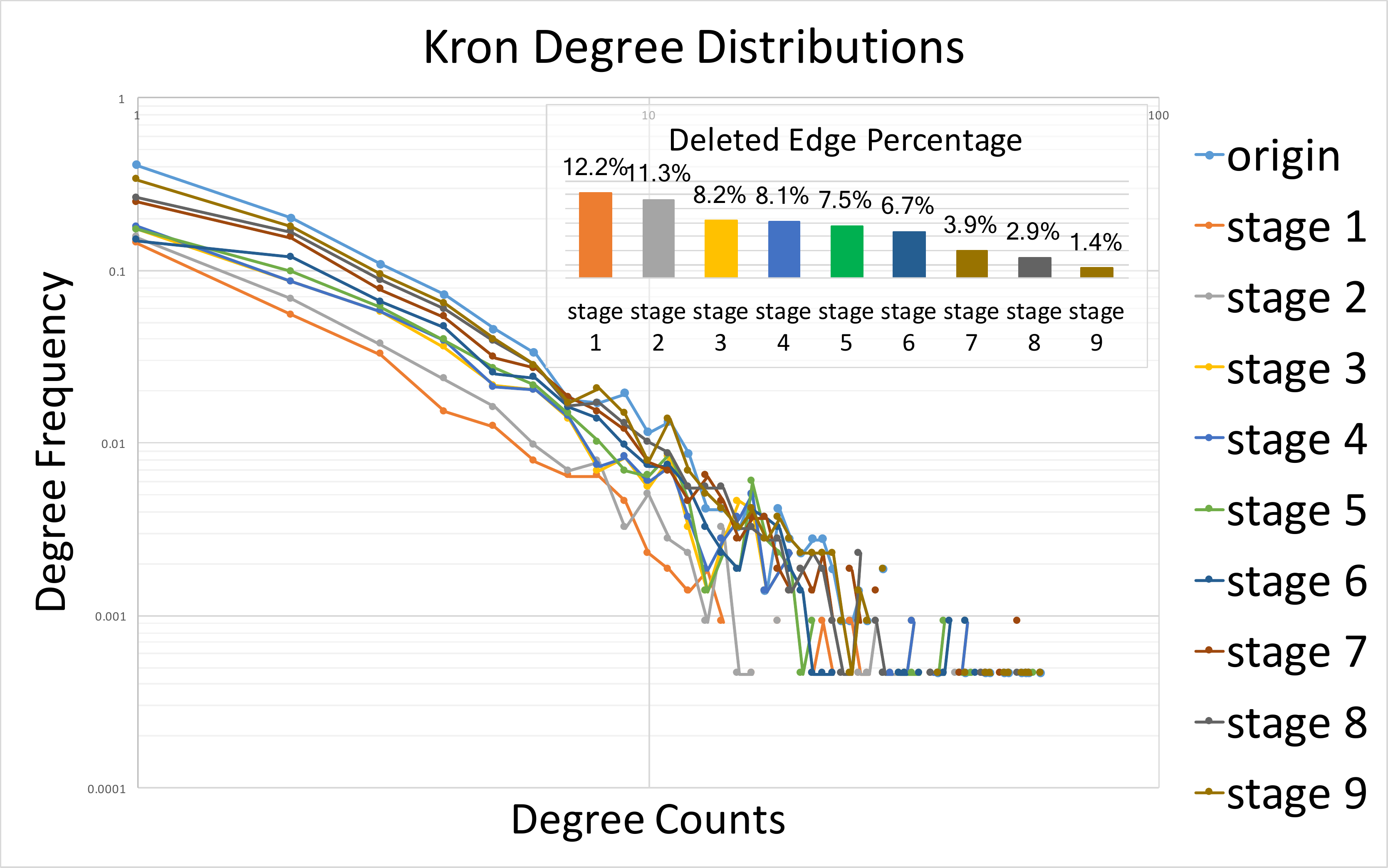}
  }
    \subfigure[Degree distributions for Facebook and wiki-vote real datasets]
  {
  	\label{real}
  \includegraphics[width=0.48\linewidth]{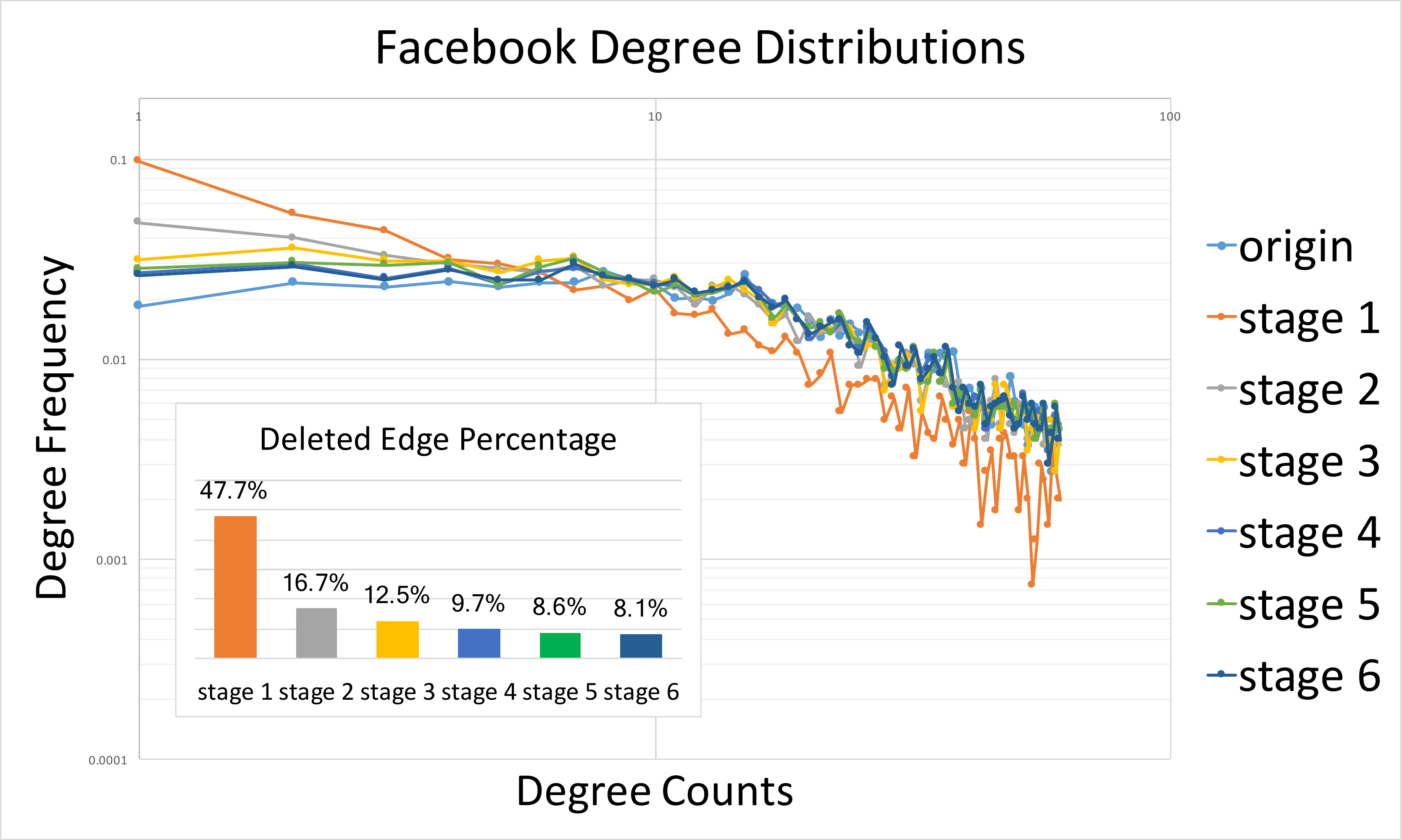}
  \includegraphics[width=0.48\linewidth]{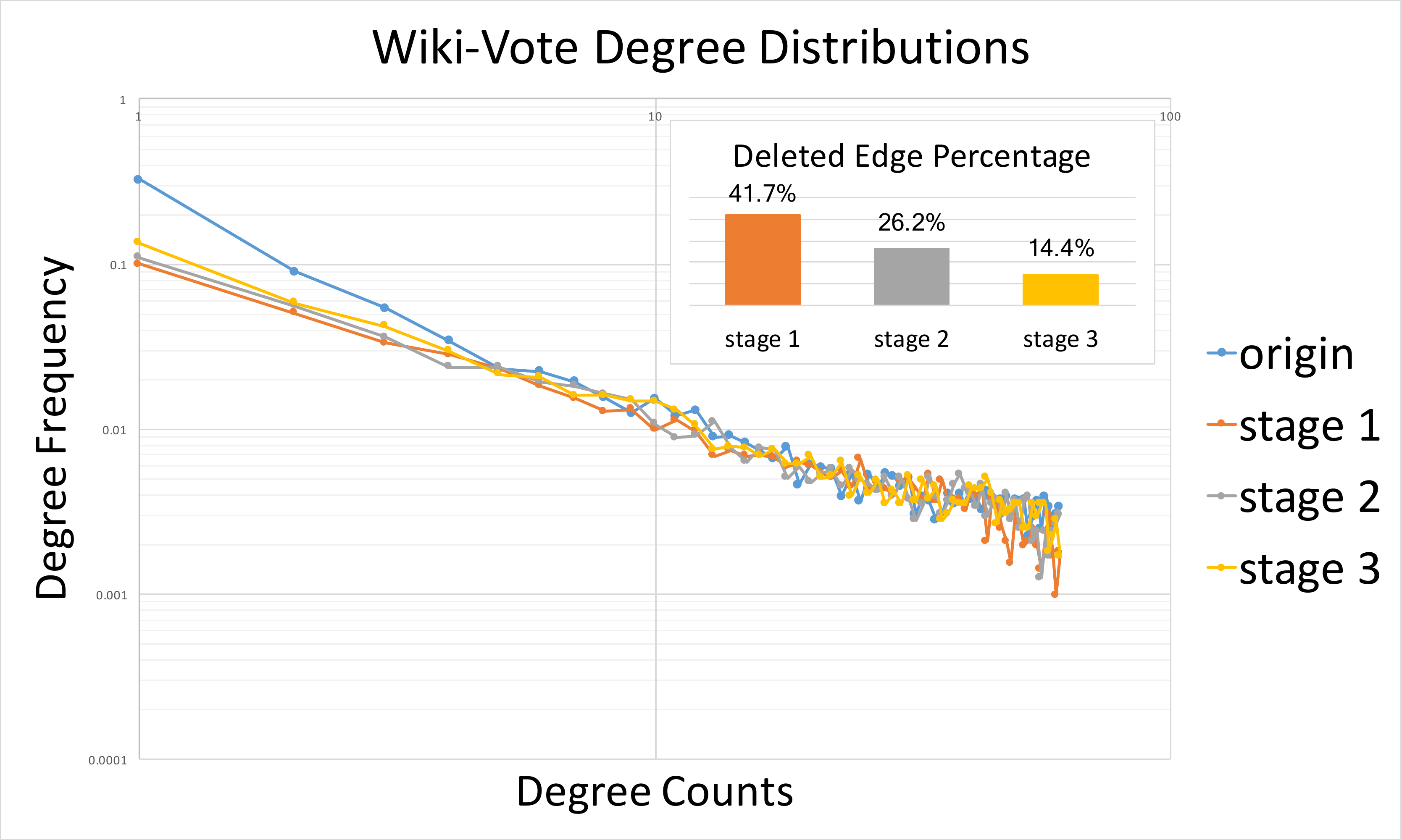}
}
  \caption{Stages and related network degree distributions for 6 datasets.}
  \label{degree-distribution}
\end{figure}

Here we demonstrate the ability of GTI reconstruction stages to preserve global topological features, which a particular focus on degree distribution.
Figure \ref{syn1}, \ref{syn2} and \ref{real} in Figure \ref{degree-distribution} shows the typical log-log degree distributions for each of the datasets given in Table \ref{tab:Table-datasets}. The horizontal axis in each degree distribution represents the number of nodes arranged, with the vertical axis representing the frequency of each degree. The blue line is used to demonstrate the degree distribution of the original graph, with other colored lines corresponding to each reconstruction stage.
In addition, for each stage in each degree distribution, we also show the ``Deleted Edge Percentage,'' which gives how many edges have been deleted in the current stage (relative to the original graph).
It can be seen that as additional stages are added (and the Deleted Edge Percentage correspondingly declines), the degree distribution becomes closer to the original network topology. What's more, we observe that practically every reconstruction stage replicates the degree distributions. Only for ER network, the first three stages learned by GTI, 95.7\% of the edges are deleted, which leads to the resulting topology of the stage cannot restore the original curve, but it is still able to reproduce the peak-like feature in the original graphs degree distribution.

\subsection{Local Topological Features} \label{local}
\begin{figure}[ht]
  \subfigure[WS Networks]
  {
  	\label{fig-WS}
    \includegraphics[width=\linewidth]{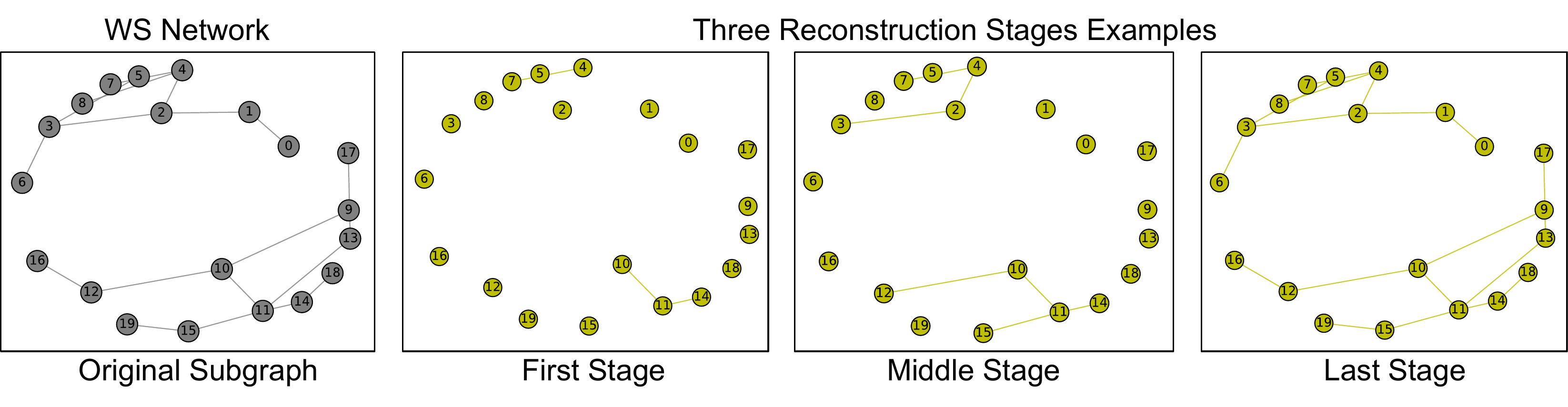}
  }
    \subfigure[Kron Networks]
  {
  	\label{fig-Kron}
    \includegraphics[width=\linewidth]{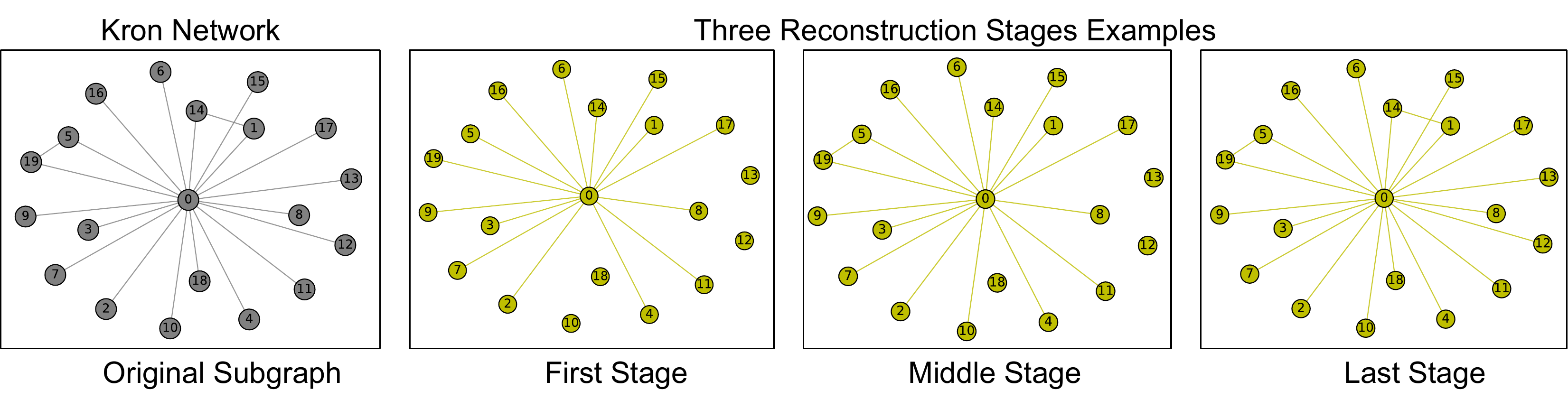}
  }
  \subfigure[Wiki-Vote Networks]
  {
  	\label{fig-wiki}
  	\includegraphics[width=\linewidth]{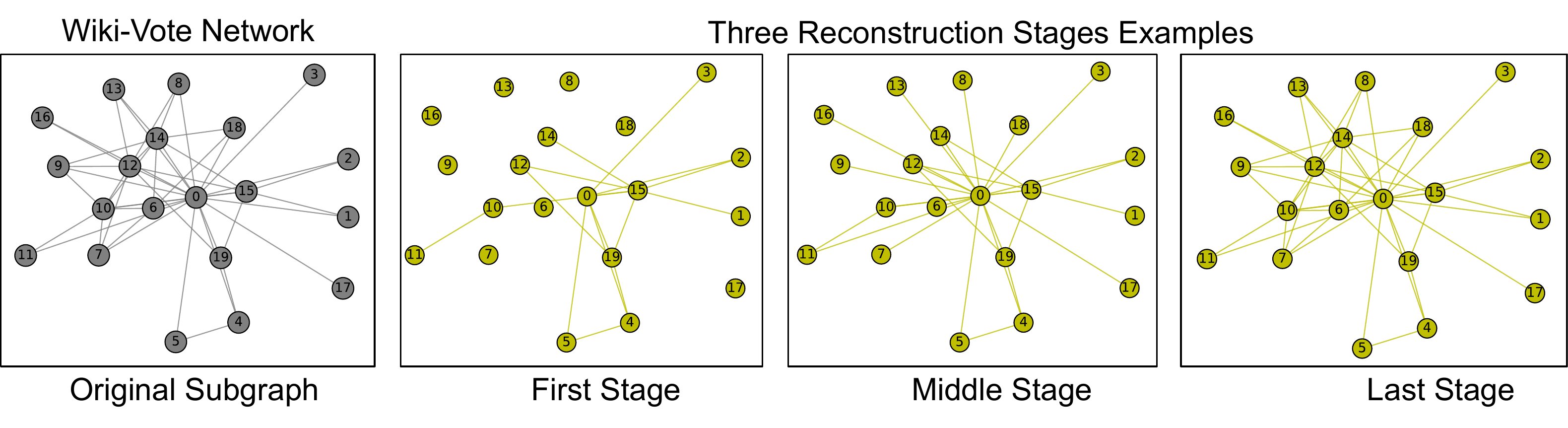}
 }
   \subfigure[Facebook Networks]
  {
  	\label{fig-facebook}
  	\includegraphics[width=\linewidth]{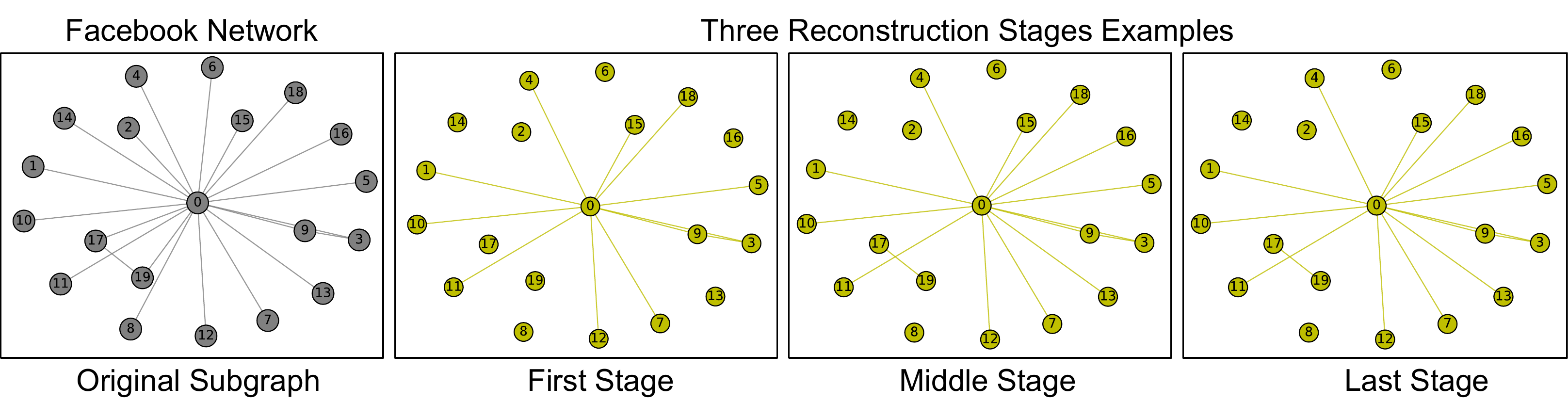}
 }
  \caption{Original subgraphs and its related stages example for four datasets.}
  \label{local-subgraph}
\end{figure}

Here we demonstrate the ability of reconstruction stages from GTI for preserving local topological features, using subgraphs with 20 nodes from two synthetic graphs and two real networks as examples. Figure \ref{local-subgraph} shows the results. The gray networks represents original subgraphs, and three yellow subgraphs shows three stages (First, Middle and Last) of the corresponding network. All of these subgraphs are drawn by Fruchterman-Reingold force-directed layout algorithm\cite{kobourov2012spring}. 

For WS network in Figure \ref{fig-WS}, as node $4$ and node $11$ are two nodes with biggest degree values. First stage and middle stage firstly reconstruct the nearby nodes of these two nodes. Then for the last stage, it reconstructs all the topology of the original subgraph.
For Kron network in Figure \ref{fig-Kron}, even in the first stage, it has already captured the star-like topological features of the original subgraph. For the last two stages, it reconstructs the full structure of the original subgraph. We argue that this phenomena is a clearly proof that the stages from GTI can be used as a indicator in demonstrating which edges are most important to the whole structure.
For Wiki-vote network in Figure \ref{fig-wiki}, the original subgraph shows that node $0$ has a largest number of neighbors. For reconstruction stages, we observe that the first stage successfully identifies node $0$, and in this stage, node $0$ also serves as the key topological structure of the entire subgraph. 
For Facebook network in Figure \ref{fig-facebook}, it is clearly to see that the first stage has successfully capture the star-like structure of node $0$, and the rest stages from GTI have retained most of the edges of the original subgraph. Since the focus of our attention is to use stages to identify important topological features of a graph, this example still shows that GTI has a good performance on capturing topologies.

\subsection{Comparison with Graph Sampling} \label{graph-sampling}
\begin{figure}[ht]
  \includegraphics[width=\linewidth]{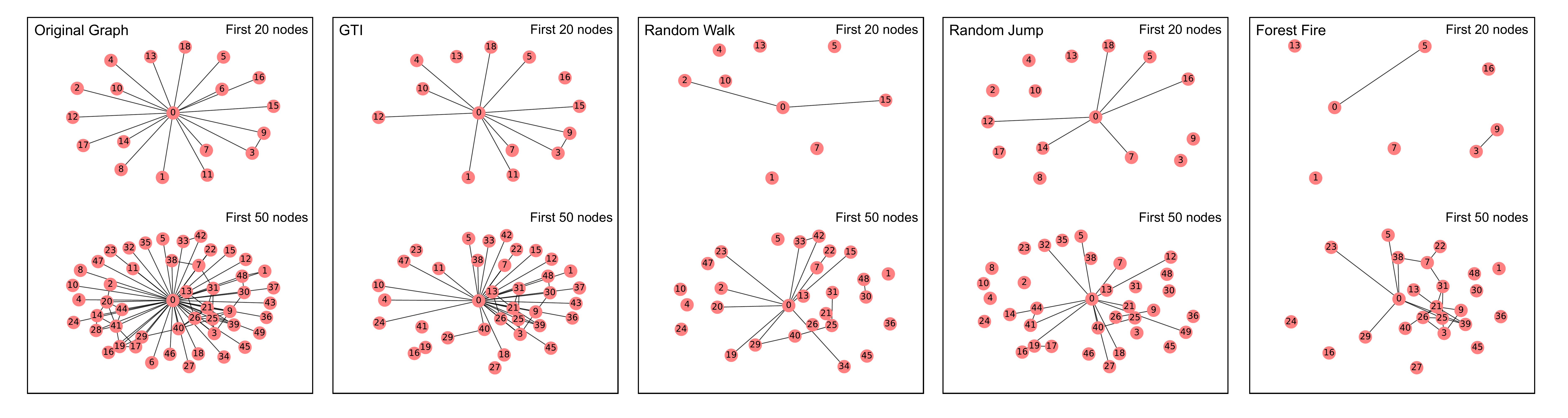}
  \caption{Comparison with graph sampling methods on the Facebook subgraphs.}
  \label{fig-sampling2}
\end{figure}

As stages in GTI can be considered as samples of the original graph, we compare the performance of GTI with three widely used graph sampling algorithms (Random Walk, Random Jump and Forest Fire \cite{leskovec2005graphs,leskovec2006sampling}) on the Facebook dataset. 
In particular, we use two subgraphs of the Facebook network (nodes 0-19 and nodes 0-49) to visually compare the ability of stage 1 of GTI to retain topological features in comparison to the three graph sampling methods\footnote{These graph sampling methods are designed to terminate with the same number of nodes as the GTI stage.}.
Figure \ref{fig-sampling2} shows the results.

One of the primary goals of graph sampling is that the sampled graph also has the ability to capture the topology of the original Graph\cite{Hu2013A}.
Through visual comparison, we observe that stage 1 of GTI has retained a similar amount of structure in the 20 and 50 node Facebook subgraphs, while either Random Walk, Random Jump or Forest Fire fails to capture the obvious star-like structure neither in 20 nodes subgraph nor 50 nodes subgraph.
In addition, as Random Walk and Random Jump have a local bias, they struggle with traversing clusters. In contrast, GTI learns very quickly about the existence of each cluster. Of course, one can improve the performance of the graph sampling methods by initializing multiple chains across all clusters, but this requires knowledge of the graph structure. This is not required by GTI, as it is an unsupervised learning method.

\section{Conclusion}
In this paper we demonstrated the ability of GANs to identify ordered stages that preserve important topological features from any arbitrary graph, and to indicate the topology reconstruction process. To the best of the authors' knowledge, this is the first paper to use GANs in such a manner. 

\bibliography{Liu}
\bibliographystyle{icml2017}

\end{document}